\documentclass{article}

\usepackage{arxiv}
\usepackage[utf8]{inputenc}
\usepackage[T1]{fontenc}
\usepackage[authoryear,round]{natbib}
\usepackage{hyperref}
\usepackage{url}
\usepackage{booktabs}
\usepackage{tabularx}
\usepackage{amsmath}
\usepackage{amsfonts}
\usepackage{amssymb}
\usepackage{microtype}
\usepackage{graphicx}
\usepackage{capt-of}
\usepackage[table]{xcolor}
\usepackage{tcolorbox}

\definecolor{FrontmatterBackground}{HTML}{F0F5F3}
\definecolor{FrontmatterInk}{HTML}{183D37}

\newtcolorbox{frontmattercard}{
    colback=FrontmatterBackground,
    colframe=FrontmatterBackground,
    boxrule=0pt,
    arc=8pt,
    outer arc=8pt,
    boxsep=0pt,
    left=14pt,
    right=14pt,
    top=14pt,
    bottom=14pt,
    before skip=0pt,
    after skip=12pt,
    width=\linewidth
}

\newcommand{\LVmonogram}{%
    \raisebox{-0.18em}{\includegraphics[height=1.15em]{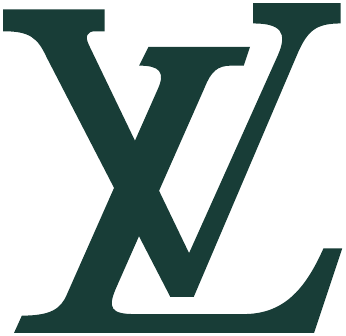}}%
}
\newcommand{\LVMHwordmark}{%
    \LVmonogram\kern0.02em MH%
}

\makeatletter
\renewcommand{\maketitle}{%
    \thispagestyle{empty}%
    {\raggedright\sffamily\bfseries\fontsize{18}{22}\selectfont
        \color{FrontmatterInk}\@title\par}%
    \vspace{9pt}%
    {\raggedright\rmfamily\bfseries\fontsize{9.3}{12}\selectfont
        \@author\par}%
    \vspace{7pt}%
    \paperAffiliations
}
\makeatother

\renewenvironment{abstract}
    {\par\vspace{10pt}\normalfont\normalsize\color{black}%
     \setlength{\parskip}{0pt}\noindent\ignorespaces}
    {\par}

\graphicspath{{Figures/}}

\newcommand{\method}{LVMH}
\newcommand{\methodlong}{Long Video Meta-memory Harness}
\newcommand{\memorymethod}{Self-Evolving Evidence Memory}
\newcommand{\memoryshort}{SEEM}
\newcommand{\frontier}{\mathcal{F}}
\newcommand{\candidate}{\mathcal{C}}
\newcommand{\dataset}{\mathcal{D}}
\newcommand{\budget}{\mathcal{B}}

\newcommand{\history}{\mathcal{H}}
\newcommand{\artifacts}{\mathcal{X}}

\title{\texorpdfstring{\LVMHwordmark}{LVMH}: A Meta-Optimization Harness for\\
Long-Video Agent Memory}

\author{%
    \bfseries
    \href{https://openreview.net/profile?id=~Liwei_Che1}{Liwei Che}\textsuperscript{1}\quad
    \href{https://openreview.net/profile?id=~Benlin_Liu1}{Benlin Liu}\textsuperscript{2}\quad
    \href{https://openreview.net/profile?id=~Dai_Li1}{Dai Li}\textsuperscript{3}\quad
    \href{https://openreview.net/profile?id=~Xiangyan_Liu1}{Xiangyan Liu}\textsuperscript{4}\quad
    \href{https://openreview.net/profile?id=~Zijian_Wu1}{Zijian Wu}\textsuperscript{4}\\
    \bfseries
    \href{https://openreview.net/profile?id=~Changjiang_Li1}{Changjiang Li}\textsuperscript{5}\quad
    \href{https://openreview.net/profile?id=~Minghao_Guo3}{Minghao Guo}\textsuperscript{1}\quad
    \href{https://openreview.net/profile?id=~Chengzhi_Mao2}{Chengzhi Mao}\textsuperscript{1}\quad
    \href{https://openreview.net/profile?id=~Vladimir_Pavlovic2}{Vladimir Pavlovic}\textsuperscript{1}\quad
    \href{https://openreview.net/profile?id=~Ranjay_Krishna1}{Ranjay Krishna}\textsuperscript{2}
}
\newcommand{\paperAffiliations}{%
    {\raggedright\normalfont\fontsize{9.3}{12}\selectfont
    \noindent
    \textsuperscript{1}\,\raisebox{-1.4pt}{\includegraphics[height=8.5pt]{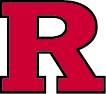}}\, Rutgers University
    \quad
    \textsuperscript{2}\,\raisebox{-1.4pt}{\includegraphics[height=8.5pt]{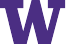}}\, University of Washington
    \quad
    \textsuperscript{3}\,\raisebox{-1.0pt}{\includegraphics[height=7.5pt]{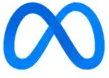}}\, Meta\\[3pt]
    \textsuperscript{4}\,\raisebox{-1.8pt}{\includegraphics[height=9.3pt]{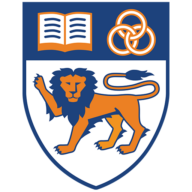}}\, National University of Singapore
    \quad
    \textsuperscript{5}\,\raisebox{-1.4pt}{\includegraphics[height=8.5pt]{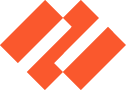}}\, Palo Alto Networks\par}
}
\date{}
\renewcommand{\shorttitle}{LVMH: Long-Video Agent Memory}

\hypersetup{
hidelinks,
pdftitle={LVMH: A Meta-Optimization Harness for Long-Video Agent Memory},
pdfsubject={Long Video Meta-Memory Harness, Multimodal Agent Memory, Long Video Understanding, Self-Evolving Memory},
pdfauthor={Liwei Che, Benlin Liu, Dai Li, Xiangyan Liu, Zijian Wu, Changjiang Li, Minghao Guo, Chengzhi Mao, Vladimir Pavlovic, Ranjay Krishna},
pdfkeywords={long video meta-memory harness, multimodal memory, long video understanding, agent memory, context management, multi-agent systems, self-evolving memory},
}

\begin{document}
\begin{frontmattercard}
\maketitle

\begin{abstract}
Long-video agents must map hours or days of multimodal evidence into a bounded context while respecting token and latency budgets, making memory design a context-management challenge.
Existing memory systems largely rely on hand-designed policies, leaving open how to jointly optimize memory design for performance and efficiency.
We propose \methodlong{} (\method{}), an automated multi-agent framework that formulates memory-policy development as multi-objective code search around a
fixed answer model.
To guard against self-certification and development-set overfitting, \method{} separates implementation from independent review and enforces protected observation boundaries.
Persistent meta-memory records past proposals, traces, costs, and failures to guide subsequent policy revisions.
Through this harness, we discover \memorymethod{} (\memoryshort{}), a ``text-first, vision-second'' policy that answers from compact structured event memory and evidence specialists, consulting localized frames through a selective visual gate.
The policy evolved on EgoLifeQA generalizes to unseen benchmarks, outperforming Meta-Harness on both Ego-R1 Bench and Video-MME (L) with GPT-5 mini.
Across the three benchmarks, \memoryshort{} with GPT-5 achieves $65.7\%$ average accuracy, surpassing the previous state of the art by $7.7$ percentage points with approximately $32\times$ lower average evaluation token use and $91\times$ shorter average response time.
\end{abstract}
\end{frontmattercard}

\begin{figure}[htbp]
    \centering
    \includegraphics[width=\linewidth]{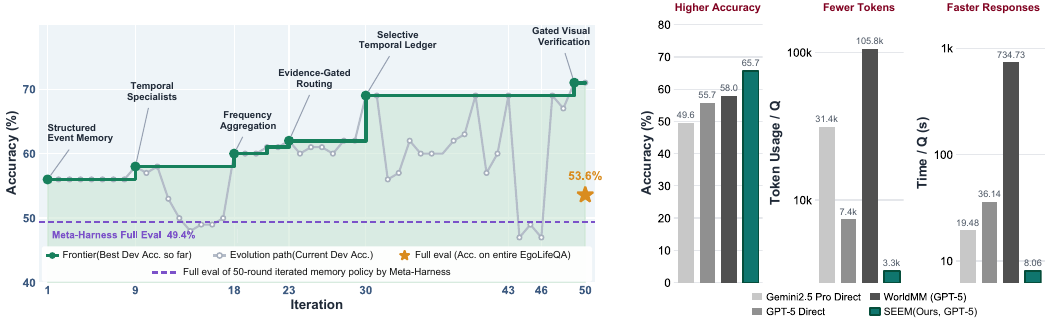}
    \caption{
    \textbf{Left:} The memory-policy frontier on the development set (an EgoLifeQA~\citep{yang2026egolifeegocentriclifeassistant} subset), annotated with introduced mechanisms. After $50$ iterations each, \method{} generalized better than Meta-Harness~\citep{lee2026metaharnessendtoendoptimizationmodel}.
    \textbf{Right:} \memoryshort{}, the final policy produced by \method{}, surpassed direct-answering and previous state-of-the-art baselines in average accuracy and efficiency across the evaluation benchmarks.}
    \label{fig:teaser}
\end{figure}

\section{Introduction}
\label{sec:introduction}

Multimodal large language models (MLLMs)~\citep{bai2025qwen3,wang2025internvl3,deepmindGemma4,ataallah2024minigpt4}
are making long-video agents increasingly practical for robotics, autonomous
driving, and video archives.
These agents must find and connect evidence across recordings that span hours
or days, even when the relevant event occupies only a few seconds.
Recordings can exceed the model's context capacity, so reasoning requires
retrieving relevant evidence at low cost.

Existing systems address this challenge with external multimodal memories that
store captions, transcripts, visual embeddings, event summaries, or
graph-structured records and retrieve a subset at answer
time~\citep{wang2024videoagentlongformvideounderstanding,fan2024videoagentmemoryaugmentedmultimodalagent,yeo2026worldmmdynamicmultimodalmemory,li2026bridgingmodalitiesspanningtime,yin2026videoarmagenticreasoninghierarchical}.
Although promising, their memory architectures and retrieval policies remain
largely designed by hand.
Human-inspired categories, such as episodic and semantic
memory~\citep{yeo2026worldmmdynamicmultimodalmemory,long2025seeinglisteningrememberingreasoning,fan2024videoagentmemoryaugmentedmultimodalagent},
need not best match an AI agent's evidence needs.
Moreover, improving accuracy together with token efficiency and latency requires
coordinated changes to storage, retrieval, and evidence presentation, making
manual design difficult.
Finally, fixed memory structures limit an agent's ability to iteratively
improve its memory policy using evaluation results.

These limitations motivate optimizing \emph{context management itself}.
A memory policy determines what to store, how to retrieve it, and when to inspect
raw visual frames before supplying evidence to an answer model.
We formulate its design as a \emph{multi-objective optimization} problem over
executable policies, jointly targeting higher accuracy, lower token use, and
shorter latency.
With the answer model fixed, the search can revise memory schemas, retrieval
rules, and visual routing based on measured performance.
Evaluation results thus guide revisions to the code that constructs the context,
with each revision assessed for effectiveness and efficiency.

Automating this search requires reliable evaluation and controlled access to
feedback.
Recent harnesses demonstrate how agents can revise memory designs through
code changes and execution
results~\citep{zhang2025memevolve,liu2026evolvememselfevolvingmemoryarchitectureautoresearch}.
Two risks become important when applying this approach to long-video memory.
\emph{Self-certification} arises when the agent that implements a candidate
also decides whether to accept it: apparent gains may hide evaluation errors
or regressions.
\emph{Development-set overfitting} arises when repeated access to individual
examples and detailed feedback encourages question-specific fixes that do not
generalize.
An effective search process therefore needs to define both who validates a
candidate and what information its developer can observe.

We introduce \methodlong{} (\method{}), a multi-agent framework that combines
policy search, independent review, and persistent \emph{meta-memory}.
Worker agents propose bounded changes and implement candidate policies in
isolated workspaces.
An orchestrator independently evaluates their accuracy, cost, temporal
validity, and regressions before deciding which candidates to promote.
Workers receive development inputs and permitted execution traces, while gold
answers, rationales, oracle timestamps, and held-out examples remain outside
their observation boundary.
Meta-memory records proposals, evaluation results, costs, failures, and
promotion decisions across iterations.
This evidence guides later proposals, helps avoid repeated failures,
and supports combining useful components from earlier candidates.
Meta-memory thus guides how policies evolve, while video memory supplies the
evidence used to answer queries.

The policy discovered through this process is \memorymethod{}
(\memoryshort{}), which follows a text-first, vision-second strategy.
It answers from compact structured event memory and uses specialists for
operations such as time-delta computation and repeated-event aggregation.
Figure~\ref{fig:teaser} (left) shows how successive policy revisions improve
development accuracy.
After $50$ iterations on a development set from EgoLife, the resulting policy achieves $53.6\%$ accuracy on the
full EgoLifeQA set with GPT-5 mini~\citep{openai2026gpt5}, beating $49.4\%$ for Meta-Harness.
Results on the held-out test split and transfer benchmarks provide further
evidence of the policy's effectiveness beyond development data.
Figure~\ref{fig:teaser} (right) shows joint gains in effectiveness and efficiency:
with GPT-5, \memoryshort{} achieves higher average accuracy across the
evaluation benchmarks, surpassing the previous state of the art with fewer tokens and lower latency.

In summary, our contributions are:
\begin{itemize}
\item We bring meta-optimization with agentic harnesses to long-video agent
memory, enabling agent-native memory policies that evolve through evaluation results
to jointly improve accuracy, token efficiency, and latency.

\item We design \methodlong{} (\method{}), a multi-agent framework intended to limit
self-certification and development-set overfitting.
Its evolved policy achieves higher validation and test accuracy than
Meta-Harness in the reported run.

\item Through \method{}, we discover \memorymethod{} (\memoryshort{}), a
memory policy that combines strong long-video reasoning with efficient
inference. It surpasses the previous state of the art in accuracy
across three benchmarks, while using fewer evaluation tokens and achieving
lower latency.
\end{itemize}

\section{Related Work}
\label{sec:related}

\noindent\textbf{Long Video Understanding.}
Long-video benchmarks require finding sparse events and relating evidence across
extended recordings~\citep{mangalam2023egoschemadiagnosticbenchmarklongform,fu2025videommefirstevercomprehensiveevaluation,wu2024longvideobenchbenchmarklongcontextinterleaved,yang2026egolifeegocentriclifeassistant}.
Much prior work focuses on improving the model through efficient architectures
and long-video training~\citep{li2026videochat}, or post-training for temporal
grounding and tool use~\citep{wang2026time,tian2025ego}.
Retrieval and video agents further enable dynamic evidence acquisition at
inference time~\citep{luo2026video,wang2024videoagentlongformvideounderstanding,fan2024videoagentmemoryaugmentedmultimodalagent}.
We optimize the memory policy that stores, selects, and presents evidence
while keeping the answer model fixed.

\noindent\textbf{Agent Memory.}
Agent memory retains information beyond the immediate context and retrieves it
for later decisions~\citep{packer2024memgptllmsoperatingsystems,sumers2024cognitivearchitectureslanguageagents,zhang2024surveymemorymechanismlarge}.
Multimodal systems extend this role to audiovisual experiences and spatial
records~\citep{long2025seeinglisteningrememberingreasoning,zou2025m33dspatialmultimodalmemory,liu2025memversemultimodalmemorylifelong},
while structured and hierarchical video memories preserve events and their
relations~\citep{li2026bridgingmodalitiesspanningtime,yin2026videoarmagenticreasoninghierarchical,yeo2025gcagentlongvideounderstandingschematic}.
Many such systems adopt human-inspired categories, such as episodic and
semantic memory~\citep{long2025seeinglisteningrememberingreasoning,yeo2026worldmmdynamicmultimodalmemory}.
Although memory contents and retrieval can adapt, their architectures remain
largely predefined.
We instead iteratively redesign the memory policy through code search,
jointly targeting accuracy, token use, and latency.

\noindent\textbf{Meta Harness Optimization.}
Meta-Harness uses execution feedback to optimize the code surrounding a
model~\citep{lee2026metaharnessendtoendoptimizationmodel}, while MemEvolve and
EvolveMem focus on evolving memory for textual agent traces and
conversations~\citep{zhang2025memevolve,liu2026evolvememselfevolvingmemoryarchitectureautoresearch}.
These approaches allow full access to all benchmark examples and evaluation feedback during search, which raises concerns about \emph{development-set overfitting}.
\method{} is designed to limit this risk by separating development, validation, and test sets for policy search and independent frontier promotion.

\section{Problem Formulation}
\label{sec:problem-formulation}
\label{background}

\noindent\textbf{Long Video Understanding.}
We study long-video understanding through question answering over recordings
spanning hours or days.
A query may require locating a brief event or connecting evidence across
distant moments, while the available frames, captions, and transcripts can
exceed the answer model's context capacity.
Given a query $q$, answer space $A_q$, and video artifacts $\artifacts_q$,
a memory policy $\mu$ constructs a compact evidence context $c_q$ for a
fixed answer model $f$:
\begin{equation}
    c_q=\mu(\artifacts_q,q,A_q;\budget),
    \qquad \hat y_q=f(q,A_q,c_q).
    \label{eq:answer-time-policy}
\end{equation}
The policy governs what to store, how to retrieve relevant evidence, and
how to assemble it under context and cost budgets $\budget$.
For a query at time $t_q$, $\artifacts_q$ contains only evidence available
by that time to preserve temporal causality.

\noindent\textbf{Memory-Policy Optimization.}
We treat context management as multi-objective search over executable
policies, targeting higher accuracy, lower token use, and shorter latency
while accounting for memory construction and answer-time costs.
With $f$ fixed, search can revise memory representations, retrieval rules,
and visual evidence selection.
At iteration $k$, candidate policies $\mu\in\candidate_k$ are evaluated on
the development set $\dataset$, yielding accuracy and cost vectors $J(\mu;\dataset)$
and execution traces.
The review operator $U$ updates the policy frontier $\frontier_k$ and persistent
meta-memory $\history_k$:
\begin{equation}
    (\frontier_{k+1},\history_{k+1})
    =U\!\left(\frontier_k,\history_k,
      \{(\mu,J(\mu;\dataset),\mathrm{trace}(\mu))\}_{\mu\in\candidate_k}\right).
    \label{eq:memory-policy-update}
\end{equation}
The frontier retains accepted policies, while meta-memory records proposals,
outcomes, failures, and review decisions to guide subsequent search.
This meta-memory supports policy development; the video memory governed by
$\mu$ supplies evidence for individual queries.
Section~\ref{sec:mm-harness} specifies candidate generation and review.

\section{Method}
\label{sec:method}


\subsection{\method{}: \methodlong{}}
\label{sec:mm-harness}
\label{sec:observation-boundary}

\method{} is a self-evolving agentic harness that automates research
and iterative improvement of long-video memory policies.
The harness separates answer-time memory use from policy evolution through
a two-layer architecture:
\begin{itemize}
    \item \emph{Memory Layer:} The answer-time multimodal memory policy $\mu$ that transforms raw video artifacts $\artifacts_q$ into a compact evidence context $c_q$ for the fixed answer model $f$.
    \item \emph{Meta Layer:} The autonomous research and governance system that guides hypothesis formulation, sandboxed implementation, and staged audits. It maintains persistent meta-memory $\history_k$ of proposals, execution traces, costs, and failures to guide iterations.
\end{itemize}

Policy evolution operates within a strict observation boundary to limit
data leakage and development-set overfitting.
We split the benchmark into a development set $\dataset$ for policy search, a held-out validation set $\mathcal{T}$ for frontier promotion, and a test set $\mathcal{S}$ for evaluation.
During candidate development, the available task inputs are development
queries $q$, answer spaces $A_q$, and pre-query video artifacts $\artifacts_q$,
alongside permitted runtime execution traces $\mathrm{trace}(\mu)$.
Gold answers, rationales, oracle timestamps, and held-out examples are excluded
from the development context.

\noindent\textbf{Initialization.} Before search, we initialize meta-memory $\history_0$ with four fixed sources
that remain unchanged throughout policy evolution.
\emph{Research Principles} specify multi-objective optimization of accuracy,
token use, and latency under temporal causality $t \le t_q$ and context
budgets $\budget$.
They prohibit shortcuts tailored to individual questions or datasets and
direct search toward memory designs that generalize across long-video
understanding tasks.
The \emph{Research Wiki} provides a literature review of long-video
understanding, while the \emph{Harness Constitution} defines the rules for
agent collaboration, evaluation, and shared-state updates.
The permitted \emph{development-set artifacts} ground this initial knowledge
in the video evidence available for policy development.
Subsequent iterations extend meta-memory with search records while preserving
these fixed materials.

\begin{figure}[t]
    \centering
    \includegraphics[width=\linewidth,trim=0 37 0 7,clip]{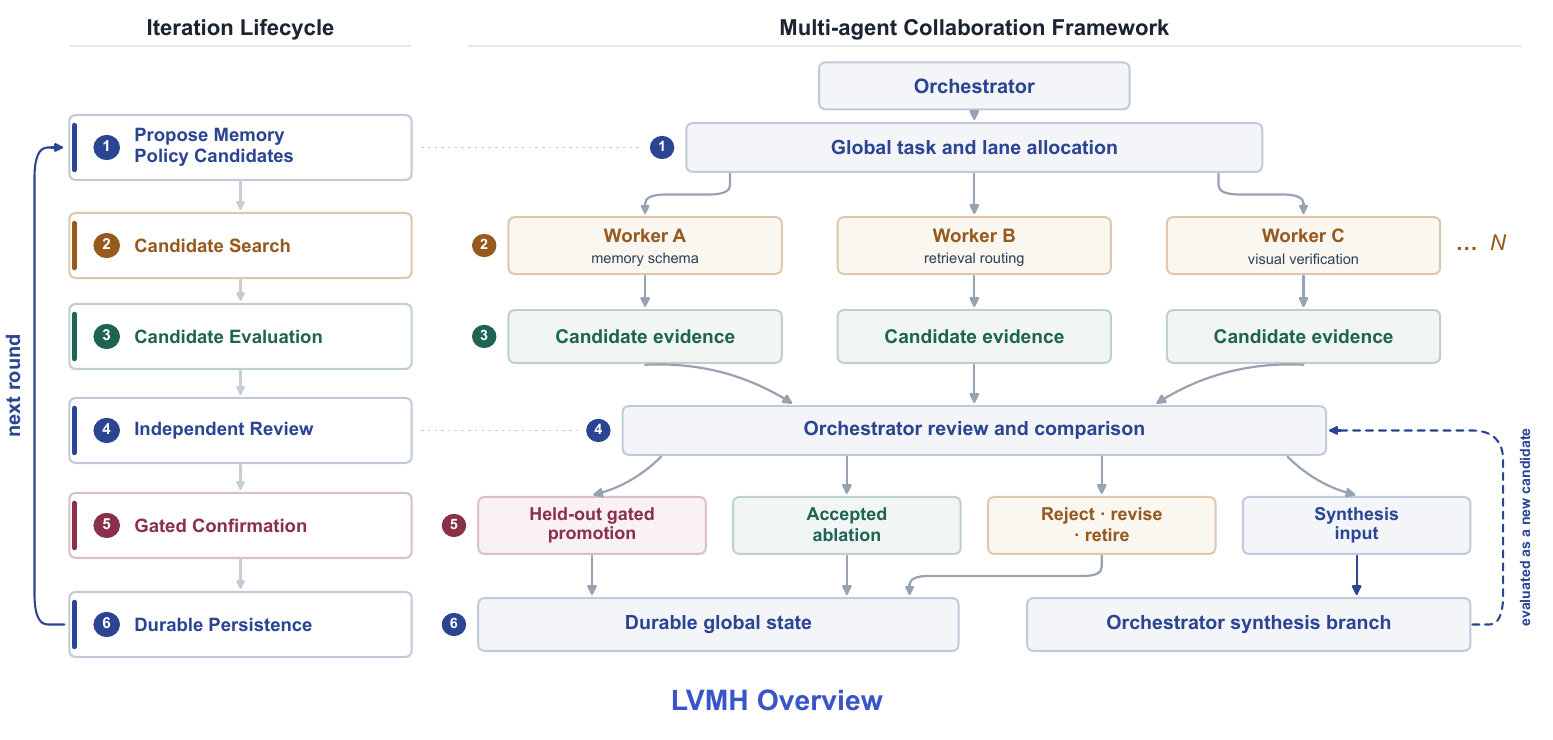}
    \caption{
    \method{} architecture and iteration lifecycle.
    \emph{Left}: Policy optimization from hypothesis proposal to meta-memory update.
    \emph{Right}: The Orchestrator assigns non-overlapping lanes; Workers develop and evaluate bounded candidates in isolated workspaces; the Orchestrator independently reviews and promotes them under controlled evaluation conditions.
    }
    \label{fig:multi-agent-self-evolution-protocol}
\end{figure}

\subsubsection{Multi-Agent Collaboration Framework}
\label{sec:iteration-lifecycle}

Within this observation boundary, the meta layer coordinates Workers and an
Orchestrator with separate responsibilities.
Workers propose, implement, and locally evaluate candidate policies
$\mu \in \candidate_k$, with read-only access to shared search state.
The Orchestrator writes no policy code: it independently reviews candidates
and alone controls held-out evaluation, frontier promotion, and updates to
meta-memory.
This separation guards against self-certification in previous autonomous
optimization frameworks~\citep{lee2026metaharnessendtoendoptimizationmodel,liu2026evolvememselfevolvingmemoryarchitectureautoresearch}.
As illustrated in Figure~\ref{fig:multi-agent-self-evolution-protocol}, each
round implements the meta-update operator $U$ through three stages:

\noindent \textbf{Hypothesis Formulation and Sandboxed Implementation.} 
Guided by the fixed initialization materials and accumulated evaluation results in
meta-memory $\history_k$, Workers pre-register proposals
$P_\mu$ specifying a bounded policy change, its expected effects on accuracy
and cost, and fallbacks.
The Orchestrator admits proposals distinct from previous attempts and assigns
them to non-overlapping search lanes.
Each Worker then implements its proposal in an isolated workspace.
Keeping hypothesis formulation and implementation with the same Worker
preserves the reasoning behind coupled changes to memory indexing, retrieval,
and context packing.

\noindent \textbf{Local Development and Independent Review.}
Workers evaluate candidates on $\dataset$, recording the evaluation vector
$J(\mu;\dataset)$ and execution traces $\mathrm{trace}(\mu)$.
They submit candidates for formal review only when results improve on the
current frontier's development performance.
The Orchestrator independently evaluates each submitted candidate against
its incumbent baseline $\mu_{\mathrm{parent}}$ using the same model $f$ and
prompt under identical cache conditions.
The audit verifies reported gains and checks temporal causality to prevent
future-event leakage.
   
\noindent \textbf{Gated Evaluation and Meta-Memory Consolidation.} 
Only candidates with independently verified development gains receive
Orchestrator-issued authorization tokens for evaluation on the held-out
validation set $\mathcal{T}$; this authorization does not expose held-out
examples to Workers.
Frontier promotion requires passing all five checks across admission,
development review, and validation:
\begin{itemize}
    \item \textbf{Novel Admission}: The proposal is bounded and structurally distinct from prior attempts ($\mu \in \candidate_k \setminus \history_k$).
    \item \textbf{Valid Execution}: The candidate yields a valid trace ($\exists\, \mathrm{trace}(\mu)$) within budget ($\text{cost}(\mu) \le \budget$), bounded by the cost of the previous state of the art~\citep{yeo2026worldmmdynamicmultimodalmemory}.
    \item \textbf{Leakage Control}: The generated context must respect temporal causality to prevent future-event leakage ($c_q = \mu(\artifacts_{\le t_q}, q, A_q; \budget)$).
    \item \textbf{Dev Dominance}: The policy improves on its parent on $\dataset$. Within the budget, an increase in cost is allowed only when the relative performance gain is at least as large as the relative cost increase; strict Pareto dominance is not required.
    \item \textbf{Transfer Audit}: The gains persist on the held-out validation set ($\mu \succ_{\mathcal{T}} \mu_{\mathrm{parent}}$).
\end{itemize}
Passing candidates enter the updated frontier $\frontier_{k+1}$.
The Orchestrator also consolidates proposals, $J(\mu;\dataset)$,
$\mathrm{trace}(\mu)$, review decisions, synthesis rules, and reflections into
$\history_{k+1}$.
This record includes attempts rejected during local development,
helping later Workers avoid repeated failures and combine complementary mechanisms.
Together, frontier promotion and meta-memory consolidation complete the
update $U$ and provide the starting state for the next round.

\subsection{\memorymethod{}}
\label{sec:self-evolving-evidence-memory}

\begin{figure}[t]
    \centering
    \includegraphics[width=\linewidth,height=0.38\textheight,keepaspectratio]{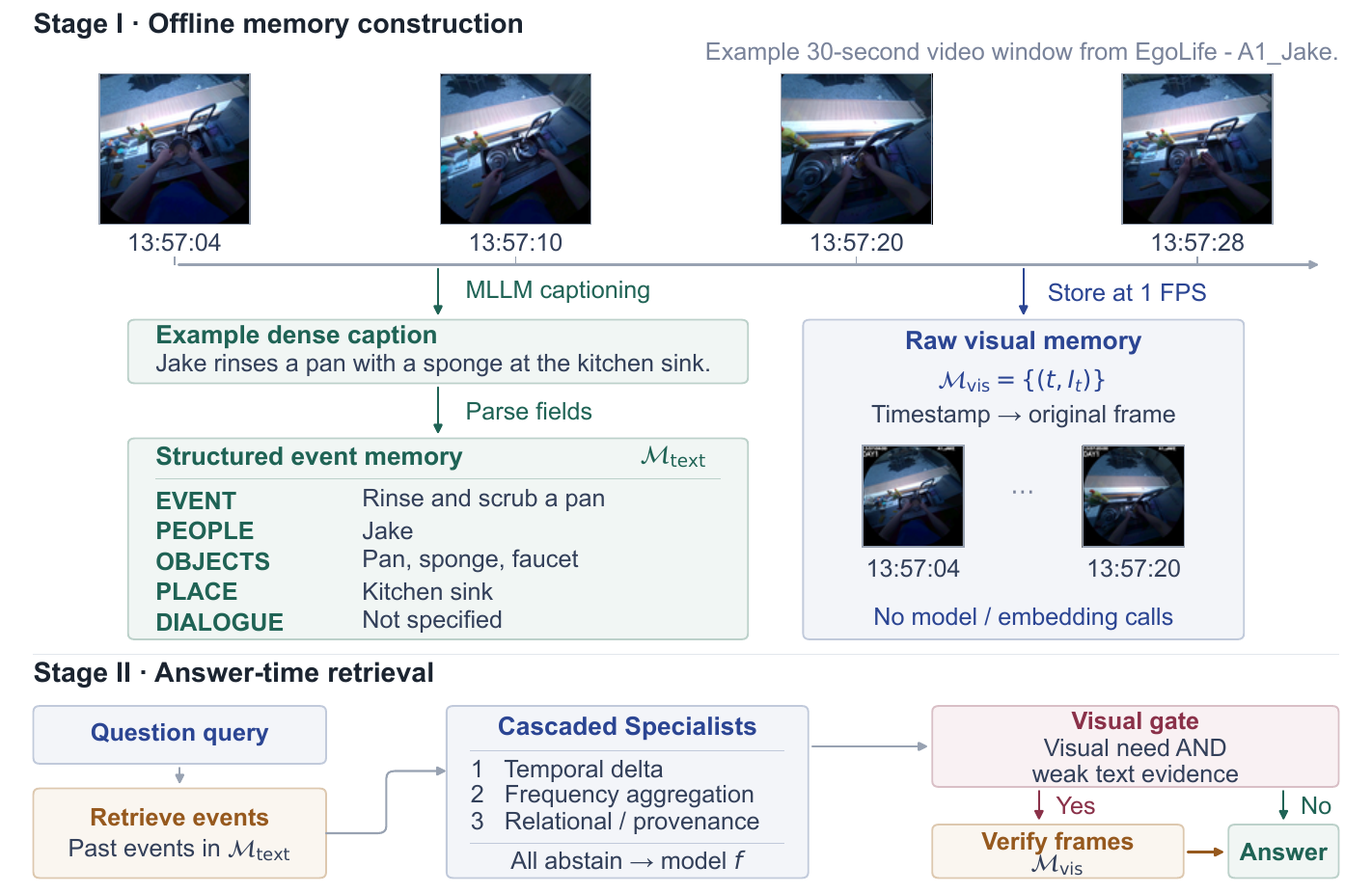}
    \caption{Overview of \memoryshort{} with an illustrative EgoLife example~\citep{yang2026egolifeegocentriclifeassistant}. Stage II uses cascaded specialists and gated visual verification.}
    \label{fig:seem_overview}
\end{figure}

\memoryshort{} is the final frontier memory policy discovered by \method{} within $50$ iterations.
\memoryshort{} adopts a ``text-first, vision-second'' strategy, answering
primarily from structured event memory and consulting raw visual frames
only when needed to verify or correct the initial answer.
Figure~\ref{fig:seem_overview} illustrates its two decoupled stages:
offline memory construction and answer-time retrieval.

\subsubsection{Stage I: Offline Memory Construction}
\label{sec:seem-stage-1}

Stage I processes each video $v$ once, constructing a structured textual event bank
$\mathcal{M}_{\text{text}}$ and a raw visual frame bank $\mathcal{M}_{\text{vis}}$:

\noindent \textbf{Structured Textual Event Bank ($\mathcal{M}_{\text{text}}$).}
The video $v$ is discretized into non-overlapping temporal windows $w_i = [t_i, t_{i} + \Delta t)$. We set $\Delta t = 30\,\text{s}$ to balance temporal granularity and context density.
For each window $w_i$, an MLLM generates a dense caption.
The caption is then parsed into a structured record $r_i$ with five semantic fields:
\begin{equation}
    r_i = \big\langle \texttt{EVENT},\, \texttt{PEOPLE},\, \texttt{OBJECTS},\, \texttt{PLACE},\, \texttt{DIALOGUE},\, w_i \big\rangle.
\end{equation}
The textual memory is defined as $\mathcal{M}_{\text{text}} = \{r_i\}_{i=1}^N$. Factorizing unstructured captions into typed fields enables fine-grained, field-aware matching during retrieval.

\noindent \textbf{Raw Visual Frame Bank ($\mathcal{M}_{\text{vis}}$).}
Instead of pre-computing dense visual embeddings or per-frame descriptions, \memoryshort{} preserves visual evidence as raw pixel arrays. 
Frames are stored by timestamp as
$\mathcal{M}_{\text{vis}} = \{(t, I_t) \mid t \in \mathcal{T}_{\text{sample}}\}$,
where $\mathcal{T}_{\text{sample}}$ denotes the set of sampled timestamps ($1$ FPS) across the video, and $I_t$ is the raw frame at time $t$. This mapping is constructed with zero model or embedding calls during offline ingestion.

\subsubsection{Stage II: Answer-Time Retrieval}
\label{sec:seem-stage-2}

At answer time, \memoryshort{} resolves query $q$ through causal filtering,
fielded retrieval, cascaded evidence specialists, and gated visual verification.

\noindent \textbf{1. Temporal Filtering and Fielded Event Retrieval.}
To preserve causal validity, retrieval uses only records available by the query time:
$\artifacts_q \triangleq \{r \in \mathcal{M}_{\text{text}} \mid r.t_{\text{end}} \le t_q\}$.
For each candidate record $r$, the system computes a field-weighted question
relevance score $S_q(r) = \sum_{\phi \in \mathcal{K}} w_\phi(q) \cdot s_\phi(r)$,
where $\mathcal{K}$ is the set of semantic fields, $w_\phi(q)$ is the
field-specific weight including question-form adjustments, and $s_\phi(r)$
denotes the lexical token overlap between the query and field $\phi$
in record $r$.
For each answer choice $c \in A_q$, a choice relevance score $S_c(r, c)$ is computed with boosted weights on exact entity and action matches. 
The total retrieval ranking score $S(r, c)$ balances individual relevance with a crucial \emph{same-record co-occurrence bonus}:
\begin{equation}
    S(r, c) = \alpha \cdot S_q(r) + \beta \cdot S_c(r, c) + \gamma \cdot \mathbb{I}\big[S_q(r) > 0 \;\wedge\; S_c(r, c) > 0\big].
    \label{eq:event-retrieval-score}
\end{equation}
This term rewards records where the question and candidate co-occur within the same event window, filtering out unrelated mentions.
Retrieval packs the top six global event anchors and top three evidence records per choice into a context buffer capped at 12,500 characters.
All scale factors are heuristically determined by the \method{} agent during
search; the final policy uses $\alpha=1$, $\beta=2$, and $\gamma=4$.
Appendix~\ref{app:fielded-retrieval} provides more details.

\noindent \textbf{2. Cascade of Evidence Specialists.}
\memoryshort{} applies three types of specialists in sequence.
\emph{Temporal delta specialists} compute elapsed time from matching events.
\emph{Frequency aggregators} count support across distinct timestamps.
\emph{Relational and provenance specialists} require queried entities to
co-occur within the same event.
Each specialist answers only when its evidence requirements are met;
otherwise, it abstains and passes the query onward.
If all abstain, the model $f$ answers from the packed event context.
The selected route produces a text answer $\hat{a}_t$ and route metadata $z_t$
for subsequent verification.
Detailed specialist rules appear in Appendix~\ref{app:specialists}.

\noindent \textbf{3. Selective Visual Escalation and Verification.}
To recover visual details, \memoryshort{} invokes a multimodal verifier only
when visual inspection is needed and text evidence is weak:
\begin{equation}
    \mathrm{OpenVisual}(q, \hat{a}_t, z_t) = \mathrm{StrongVisualNeed}(q) \;\wedge\; \mathrm{WeakText}(\hat{a}_t, z_t).
    \label{eq:visual-gate}
\end{equation}
When the gate is closed, \memoryshort{} returns the text answer directly.
When it opens, $f$ inspects timestamp-localized frames from
$\mathcal{M}_{\text{vis}}$ and is instructed to retain $\hat{a}_t$ unless
visual evidence clearly supports an alternative.
If frames are unavailable or verification fails, the text answer is preserved.
Appendix~\ref{app:visual-verification} provides details.

\section{Experiments}


\noindent\textbf{Data and Baselines.} We evaluate \memoryshort{} on three benchmarks (detailed in Appendix~\ref{app:datasets}): EgoLifeQA \citep{yang2026egolifeegocentriclifeassistant}, Ego-R1 Bench~\citep{tian2025ego}, and Video-MME \citep{fu2025videommefirstevercomprehensiveevaluation}. For \method{} and Meta-Harness, we randomly partition EgoLifeQA's $500$ questions into $100$ for policy development, $200$ for validation during frontier promotion, and $200$ for final testing. The test split is excluded from policy search and frontier promotion. Ego-R1 Bench and Video-MME (L) are used to evaluate transfer.
We compare \memoryshort{} with video LLMs~\citep{openai2026gpt5,bai2025qwen3,google2026geminipro}, long-video LLMs~\citep{li2026videochat,wang2026time,wang2025video}, RAG-based video systems~\citep{luo2026video,guo2025lightrag,gutierrez2024hipporag}, memory-based video agents~\citep{tian2025ego,long2025seeinglisteningrememberingreasoning,yeo2026worldmmdynamicmultimodalmemory}, and harness optimization~\citep{lee2026metaharnessendtoendoptimizationmodel}.

\noindent\textbf{Implementation.} We use GPT-5.5~\citep{openai2026gpt5} as the search agent for the memory evolution of \method{} and Meta-Harness. Memory construction, retrieval, and gating are implemented with GPT-5 mini~\citep{openai2026gpt5}. We evaluate the final memory policy produced by \method{}, namely \memoryshort{}. Further implementation details are provided in Appendix~\ref{app:implementation}.

\begin{table}[t]
\definecolor{developmentred}{HTML}{F08080}
\definecolor{testingblue}{HTML}{87CEEB}
\caption{Evaluation results (accuracy, \%) on long-video understanding benchmarks. \textbf{Bold} marks the best result. \protect\colorbox{developmentred!30}{Red shading} denotes results on the full 500-question EgoLifeQA set, including 100 development, 200 validation, and 200 test questions. \protect\colorbox{testingblue!30}{Blue shading} denotes transfer benchmarks that are not observable to the agent during iterative optimization.}
\label{tab:main-results}
\begin{center}
\small
\setlength{\aboverulesep}{0pt}
\setlength{\belowrulesep}{0pt}
\setlength{\extrarowheight}{1pt}
\resizebox{\linewidth}{!}{%
\begin{tabular}{llccccccccc}
\toprule
& & \multicolumn{6}{c}{EgoLifeQA} & & & \\
\cmidrule(lr){3-8}
Method & Model & Ent. & EvR. & Hab. & Rel. & Task & Avg. & Ego-R1 Bench & Video-MME (L) & Avg. \\
\midrule
\multicolumn{2}{l}{\textit{Base Models}} & & & & & & & & & \\
Qwen3-VL-8B & 8B & 35.2 & 30.2 & 39.3 & 46.4 & 46.0 & 38.6 & 35.7 & 48.0 & 40.8 \\
Gemini 2.5 Pro & -- & 43.2 & 40.5 & 41.0 & 55.2 & 52.4 & 46.4 & 46.7 & 55.7 & 49.6 \\
GPT-5 mini & -- & 30.4 & 38.1 & 37.7 & 27.2 & 36.5 & 33.2 & 34.3 & 54.1 & 40.5 \\
GPT-5 & -- & 47.2 & 42.1 & 47.5 & 53.6 & 55.6 & 48.6 & 46.3 & 72.3 & 55.7 \\
\midrule
\multicolumn{2}{l}{\textit{Long Video LLMs}} & & & & & & & & & \\
VideoChat-Flash & 7B & 28.8 & 32.5 & 37.7 & 37.6 & 38.1 & 34.2 & 42.7 & 44.1 & 40.3 \\
Time-R1 & 7B & 39.2 & 50.8 & 65.6 & 48.8 & 47.6 & 48.8 & 48.0 & 37.6 & 44.8 \\
Video-RTS & 7B & 40.8 & 48.4 & 62.3 & 48.8 & 47.6 & 48.2 & 47.4 & 47.9 & 47.8 \\
\midrule
\multicolumn{2}{l}{\textit{RAG-based Video LLMs}} & & & & & & & & & \\
LightRAG & -- & 40.8 & 48.4 & 67.2 & 50.4 & 44.4 & 48.8 & 52.3 & 46.6 & 49.2 \\
HippoRAG & -- & 48.8 & \textbf{60.3} & 70.5 & \textbf{60.8} & \textbf{66.7} & 59.6 & 56.0 & 52.1 & 55.9 \\
Video-RAG & -- & 49.6 & 56.3 & 67.2 & 55.2 & 54.0 & 55.4 & 49.7 & 55.4 & 53.5 \\
\midrule
\multicolumn{2}{l}{\textit{Memory-based Video LLMs}} & & & & & & & & & \\
EgoRAG & -- & 40.0 & 56.3 & 62.3 & 54.4 & 52.4 & 52.0 & 49.0 & 41.1 & 47.4 \\
Ego-R1 & 3B & 51.2 & 53.2 & 63.9 & 50.4 & 50.8 & 53.0 & 52.0 & 42.7 & 49.2 \\
HippoMM & -- & 45.6 & 53.2 & 70.5 & 55.2 & 58.7 & 54.6 & 53.0 & 41.6 & 49.7 \\
M3-Agent & 32B & 44.4 & 54.8 & 62.3 & 56.8 & 54.0 & 53.5 & 52.0 & 55.3 & 53.6 \\
WorldMM & GPT-5 mini & 41.6 & 42.9 & 41.0 & 44.0 & 49.2 & 43.4 & 50.3 & 52.6 & 48.8 \\
WorldMM & GPT-5 & 50.4 & 48.4 & 59.0 & 45.6 & 58.7 & 50.8 & 50.6 & 72.6 & 58.0 \\
\midrule
\multicolumn{2}{l}{\textit{Automated Harness Optimization}} & & & & & & & & & \\
Meta-Harness & GPT-5 mini & \cellcolor{developmentred!30} 45.6 & \cellcolor{developmentred!30} 50.0 & \cellcolor{developmentred!30} 52.5 & \cellcolor{developmentred!30} 52.8 & \cellcolor{developmentred!30} 46.0 & \cellcolor{developmentred!30} 49.4 & \cellcolor{testingblue!30} 42.0 & \cellcolor{testingblue!30} 47.2 & 46.2 \\
\textbf{\memoryshort{} (ours)} & GPT-5 mini & \cellcolor{developmentred!30} 47.2 & \cellcolor{developmentred!30} 55.6 & \cellcolor{developmentred!30} 65.6 & \cellcolor{developmentred!30} 51.2 & \cellcolor{developmentred!30} 55.6 & \cellcolor{developmentred!30} 53.6 & \cellcolor{testingblue!30} 53.0 & \cellcolor{testingblue!30} 55.9 & 54.2 \\
\textbf{\memoryshort{} (ours)} & Qwen3-VL-8B & \cellcolor{developmentred!30} 50.4 & \cellcolor{developmentred!30} 49.2 & \cellcolor{developmentred!30} 63.9 & \cellcolor{developmentred!30} 47.2 & \cellcolor{developmentred!30} 54.0 & \cellcolor{developmentred!30} 51.4 & \cellcolor{testingblue!30} 50.3 & \cellcolor{testingblue!30} 55.1 & 52.3 \\
\textbf{\memoryshort{} (ours)} & GPT-5 & \cellcolor{developmentred!30} \textbf{61.6} & \cellcolor{developmentred!30} 55.6 & \cellcolor{developmentred!30} \textbf{78.7} & \cellcolor{developmentred!30} 52.8 & \cellcolor{developmentred!30} \textbf{66.7} & \cellcolor{developmentred!30} \textbf{60.6} & \cellcolor{testingblue!30} \textbf{62.0} & \cellcolor{testingblue!30} \textbf{74.4} & \textbf{65.7} \\

\bottomrule
\end{tabular}}
\end{center}
\end{table}

\subsection{Evaluation}

\noindent \textbf{Main Results.}
With GPT-5, \memoryshort{} achieves the highest average accuracy in Table~\ref{tab:main-results}, exceeding WorldMM by $7.7$ percentage points.
With GPT-5 mini, \memoryshort{} reaches $54.2\%$ overall, exceeding WorldMM and Meta-Harness by $5.4$ and $8.0$ percentage points, respectively.
Compared to Meta-Harness, the gains are $4.2$ points on EgoLifeQA, $11.0$ on Ego-R1 Bench, and $8.7$ on Video-MME (L). Figure~\ref{fig:egolife-frontier-dev-vs-vali} shows that \memoryshort{} also outperforms Meta-Harness on EgoLifeQA's validation and test splits. These results indicate that \method{} limits development-set overfitting.
The policy also improves Qwen3-VL-8B from $40.8\%$ to $52.3\%$ overall, indicating that its benefits extend beyond the backbone used during policy evolution.

\noindent \textbf{Efficiency Analysis.}
Table~\ref{tab:ablation-latency} reports GPT-5 evaluation costs for \memoryshort{}, direct answering,
and WorldMM. \memoryshort{} uses fewer tokens and has lower per-question latency than both baselines, with no more model requests. On Video-MME (L), it reduces token use by $40.6\times$ and wall-clock time by $136\times$ compared to WorldMM. On EgoLifeQA, specialists reduce the average number of model requests per question below $1$. Offline construction also uses fewer tokens than WorldMM (Table~\ref{tab:memory_construction_cost}).

\begin{table}[tbp]
\begin{minipage}[c]{0.48\linewidth}
\caption{GPT-5 cost per question: requests, tokens (k), and time (s). Direct uses no memory.}
\label{tab:ablation-latency}
\centering
\begin{tabular}{@{}lrrr@{}}
\toprule
Method & Req. & Tok. & Time \\
\midrule
\multicolumn{4}{l}{\textit{EgoLifeQA}} \\
Direct & 1.00 & 11.34k & 43.07 \\
WorldMM & 30.97 & 36.21k & 140.51 \\
\memoryshort{} & \textbf{0.97} & \textbf{3.57k} & \textbf{9.76} \\
\midrule
\multicolumn{4}{l}{\textit{Video-MME (L)}} \\
Direct & \textbf{1.00} & 6.93k & 24.94 \\
WorldMM & 201.20 & 134.02k & 981.70 \\
\memoryshort{} & \textbf{1.00} & \textbf{3.30k} & \textbf{7.22} \\
\bottomrule
\end{tabular}
\end{minipage}\hfill
\begin{minipage}[c]{0.48\linewidth}
\centering
\includegraphics[width=\linewidth]{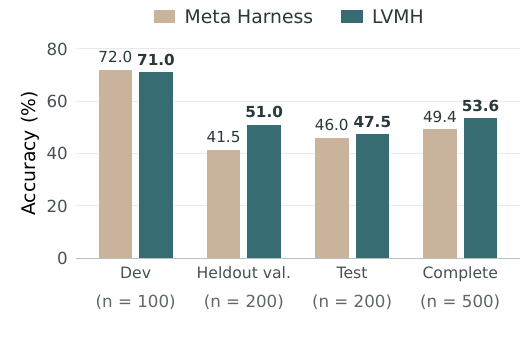}
\captionof{figure}{EgoLifeQA accuracy by split for the final frontier policies. \method{} outperforms Meta-Harness on validation and test.}
\label{fig:egolife-frontier-dev-vs-vali}
\end{minipage}
\end{table}

\subsection{Ablation}
\label{sec:ablation}

\begin{figure}[htbp]
\centering
\includegraphics[width=\linewidth]{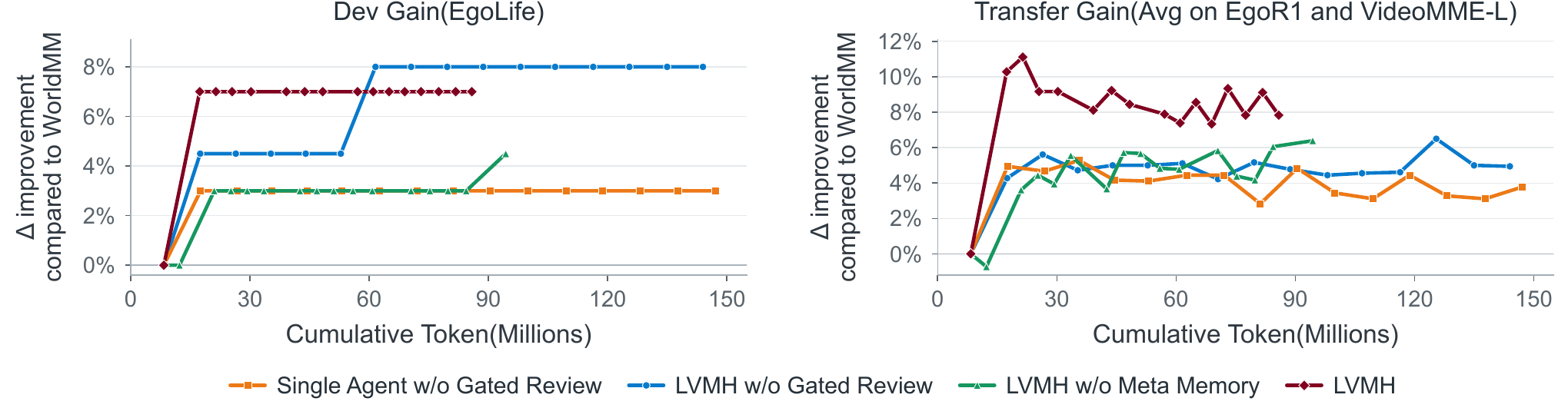}
\caption{Harness ablation: four settings run concurrently from scratch for $15$ iterations.
Curves show gains over WorldMM versus cumulative search tokens (millions).
Left: EgoLifeQA development. Right: mean transfer gain on Ego-R1 Bench and Video-MME (L).
Gated review reduces token use; meta-memory accelerates development gains.}
\label{fig:ablation-lvmh-token-budget}
\end{figure}

\noindent\textbf{Harness Design.}
Figure~\ref{fig:ablation-lvmh-token-budget} compares four settings:
the complete \method{}, \method{} without gated review,
\method{} without meta-memory, and a single-agent variant without
gated review.
We run all four settings concurrently from scratch for $15$ iterations
each and plot their gains against cumulative token consumption.
Without gated review, the agent must rely on validation evaluations
to decide whether to promote a candidate to the frontier, substantially
increasing token consumption over iterations.
Gated review reduces this cost by screening candidates before validation.
Retaining meta-memory accelerates development-set improvement.
Records of failed attempts and earlier lessons help the agent avoid
retrying unsuccessful designs and guide subsequent proposals.
The complete framework also maintains better transfer gains
and faster development progress.

\par
\noindent
\begin{minipage}[t]{0.49\linewidth}
\vspace{0pt}
\textbf{\memoryshort{} Component Ablation.}
Table~\ref{tab:ablation-seem} evaluates \memoryshort{}'s four components.
Removing structured memory causes the largest drop ($4.8$ percentage points).
Removing specialists, the co-occurrence bonus, or gated visual verification
reduces accuracy by $2.4$, $3.0$, or $2.8$ points, respectively.
Appendix~\ref{app:visual-gate-ablation} compares rule-based and LLM gates.
\end{minipage}\hfill
\begin{minipage}[t]{0.49\linewidth}
\vspace{0pt}
\setlength{\abovecaptionskip}{0pt}
\captionof{table}{Component ablation of \memoryshort{}.
Each variant removes one component.}
\label{tab:ablation-seem}
\centering
\small
\begin{tabular}{@{}l@{\hspace{6pt}}c@{}}
\toprule
Variant & Accuracy (\%) \\
\midrule
\textbf{\memoryshort{}} & \textbf{53.6} \\
w/o Structured Memory & 48.8 \\
w/o Specialists & 51.2 \\
w/o Co-occurrence Bonus & 50.6 \\
w/o Gated Visual Verification & 50.8 \\
\bottomrule
\end{tabular}
\end{minipage}
\par

\subsection{Evolution Process Analysis}
\label{sec:evolution-process-analysis}

\begin{figure}[htbp]
\centering
\includegraphics[width=\linewidth]{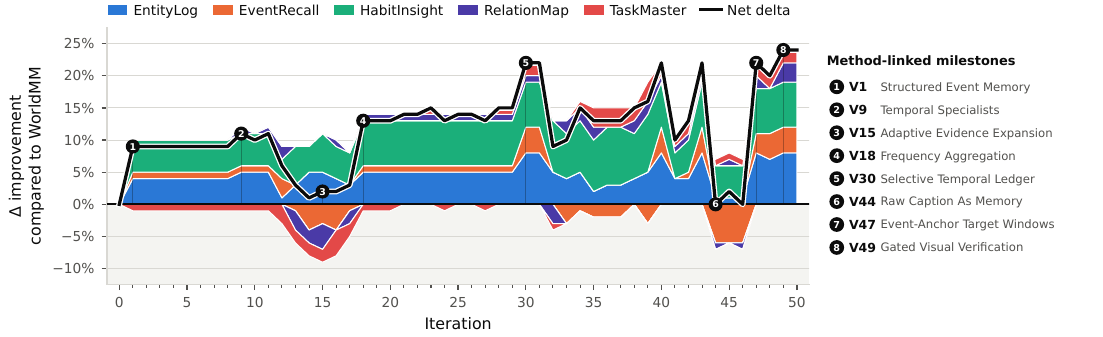}
\caption{EgoLifeQA development gains by question type, annotated with policy changes.}
\label{fig:egolife_dev_official_problem_type_delta_stacked_area_clean}
\end{figure}

Figure~\ref{fig:egolife_dev_official_problem_type_delta_stacked_area_clean}
traces successive revisions that suggest three lessons for long-video
memory design.

\noindent\textbf{Efficiency.}
The search first reduces context size through structured event memory
(V1), then reduces model calls through temporal and relational specialists
(V9) and frequency aggregation (V18).
Memory thus evolves from a compact evidence store to support explicit computation.
This suggests two routes to efficiency: compress the evidence supplied to the
model and resolve sufficiently supported queries through memory operations.

\noindent\textbf{Information Loss and Localization.}
Fixed windows and structured compression can discard details and fragment
continuous events. Broad narrative expansion (V15) reduces EventRecall,
while a selective temporal ledger (V30) better connects relevant events.
Replacing fields with raw captions (V44) again causes a regression;
expansion around retrieved event anchors (V47) recovers useful neighboring
context. Recovering information thus requires accurate localization:
preserve structured anchors, then expand around events relevant to the question.

\noindent\textbf{The Role of Visual Information.}
Text retrieval cannot recover visual details omitted during memory
construction. Gated verification (V49) adds a complementary source of
evidence, reaching the highest development accuracy in the trajectory.
Yet broader visual invocation does not improve accuracy
(Appendix~\ref{app:visual-gate-ablation}). This suggests using text memory for
reasoning across long timescales and consulting frames selectively to
verify unresolved appearance, spatial, or counting details.

\section{Conclusion}


We addressed context and budget management for long-video agents through
automated meta-optimization of multimodal memory design.
\method{}, a multi-agent harness, combines strict observation boundaries
and independent orchestrator auditing to mitigate self-certification and
development-set overfitting.
It discovered \memoryshort{}, a lightweight, ``text-first, vision-second''
memory policy that achieves superior accuracy with significantly fewer
tokens and lower latency.
Its evolution suggests three lessons for long-video agent design:
(1) memory should retrieve and distill key evidence rather than pass all
relevant information to the model;
(2) visual evidence should complement memory reasoning, with its value
balanced against processing cost; and
(3) policies based on rules and text can be efficient and effective.

\bibliography{references}
\bibliographystyle{unsrtnat}

\clearpage
\appendix
\section{Evaluation Datasets and Baselines}
\label{app:datasets}

We evaluate on two benchmarks for memory over daily egocentric recordings and
one benchmark for general video understanding.
Table~\ref{tab:dataset-overview} summarizes the benchmarks used in our experiments.

\begin{table}[htbp]
\caption{Benchmarks used in our experiments. EgoLifeQA contains the 500
questions in the public A1 Jake release. Video-MME (L) contains three questions
for each of 300 long videos.}
\label{tab:dataset-overview}
\centering
\begin{tabular}{lrl}
\toprule
Benchmark & Questions & Role in Our Experiments \\
\midrule
EgoLifeQA & 500 & 100 development / 200 validation / 200 test \\
Ego-R1 Bench & 300 & Transfer evaluation \\
Video-MME (L) & 900 & Transfer evaluation \\
\bottomrule
\end{tabular}

\end{table}

\noindent\textbf{EgoLifeQA.}
EgoLifeQA~\citep{yang2026egolifeegocentriclifeassistant} evaluates memory and
reasoning over daily life.
The EgoLife collection contains 266 hours of retained first-person audiovisual
footage from six participants living together for seven days.
The paper describes 3,000 multiple-choice questions, with 500 questions per
participant, but the public QA release contains only the 500-question
A1 Jake subset; the full 3,000-question set has not been released.
Its five categories are \emph{EntityLog} (entities and their attributes),
\emph{EventRecall} (past events), \emph{HabitInsight} (recurring behaviors),
\emph{RelationMap} (interpersonal relations), and \emph{TaskMaster}
(task-related reasoning).
Questions are anchored to a query time and may require evidence from earlier
days.
Our experiments use this entire public QA release, corresponding to
approximately 51.6 hours of Jake's video.

\noindent\textbf{Ego-R1 Bench.}
Ego-R1 Bench~\citep{tian2025ego} contains 300 multiple-choice questions,
distributed evenly across six first-person perspectives in EgoLife.
Each perspective has 50 questions drawn from human-labeled and human-verified
examples.
The benchmark emphasizes detailed visual reasoning and spatial and temporal
relations across multiple observations.
We evaluate on all 300 questions.
This benchmark is separate from Ego-R1's training data, which contain
question-answer pairs and tool-use reasoning traces.
Ego-R1 Bench evaluates transfer across benchmark questions and participant
perspectives within the same egocentric recording environment.

\noindent\textbf{Video-MME.}
Video-MME~\citep{fu2025videommefirstevercomprehensiveevaluation} contains 900
videos totaling approximately 254 hours, with 2,700 human-annotated
multiple-choice questions.
The videos cover six broad domains and 30 subfields, and the questions test
12 perception and reasoning tasks.
Video durations range from 11 seconds to one hour, with 300 videos in each
of the short, medium, and long groups.
We use the long split: 300 videos lasting 30 to 60 minutes and 900 questions.
The benchmark also provides audio tracks and, where available, subtitles.
Its diverse video sources assess transfer beyond the shared egocentric
recording environment of EgoLifeQA and Ego-R1 Bench.

\noindent\textbf{Evaluation protocol.}
We report multiple-choice accuracy, computed as the proportion of correctly
answered questions.
For both \method{} and Meta-Harness, we randomly partition the 500 Jake questions
into 100 for policy development, 200 for validation during frontier promotion,
and 200 for final testing.
The test split is excluded from policy search and frontier promotion.
The reported 500-question EgoLifeQA results include all three splits;
Figure~\ref{fig:egolife-frontier-dev-vs-vali} reports accuracy separately for each split.
Ego-R1 Bench and Video-MME (L) are used for transfer evaluation.
For timestamped egocentric questions, memory access is restricted to evidence
available by the query time (Section~\ref{sec:observation-boundary}).

\noindent\textbf{Baselines.}
Table~\ref{tab:main-results} groups the baselines into five categories:
\emph{base models}, including Qwen3-VL-8B~\citep{bai2025qwen3},
Gemini 2.5 Pro~\citep{google2026geminipro}, and GPT-5 mini and GPT-5~\citep{openai2026gpt5};
\emph{long-video LLMs}, including VideoChat-Flash~\citep{li2026videochat},
Time-R1~\citep{wang2026time}, and Video-RTS~\citep{wang2025video};
\emph{RAG-based video systems}, including LightRAG~\citep{guo2025lightrag},
HippoRAG~\citep{gutierrez2024hipporag}, and Video-RAG~\citep{luo2026video};
\emph{memory-based video agents}, including EgoRAG~\citep{yang2026egolifeegocentriclifeassistant},
Ego-R1~\citep{tian2025ego}, HippoMM~\citep{lin2026hippomm},
M3-Agent~\citep{long2025seeinglisteningrememberingreasoning}, and
WorldMM~\citep{yeo2026worldmmdynamicmultimodalmemory};
and \emph{automated harness optimization}, represented by
Meta-Harness~\citep{lee2026metaharnessendtoendoptimizationmodel}.
We reproduce the direct-evaluation results for the base models and the results
for WorldMM and Meta-Harness; all other baseline results are taken from their
respective papers.

\section{Method Details}
\label{app:implementation}
\label{app:method-details}

We detail \memoryshort{}'s temporal filtering, fielded event retrieval,
evidence specialists, and visual verification
(Section~\ref{sec:seem-stage-2}).
We retain the main-text notation: $q$ is the question, $A_q$ its answer
space, $t_q$ its observation cutoff, $c_q$ the packed evidence context,
and $f$ the fixed answer model.
An individual answer choice is denoted by $c\in A_q$.

\noindent\textbf{Meta-Harness reproduction.}
We reproduce Meta-Harness using the same GPT-5.5 search agent as \method{},
with the same budget of $50$ search iterations and identical EgoLifeQA
development, validation, and test splits ($100$/$200$/$200$ questions).
For both methods, the test split is excluded from policy search and frontier
promotion.

\subsection{Temporal Filtering and Fielded Event Retrieval}
\label{app:fielded-retrieval}

\noindent\textbf{Temporal eligibility.}
Each textual record summarizes a $30$-second event window
$[r.t_{\text{start}},r.t_{\text{end}})$ and retains five fields:
\texttt{EVENT}, \texttt{PEOPLE}, \texttt{OBJECTS}, \texttt{PLACE}, and
\texttt{DIALOGUE}.
For $r=r_i$, this is the window $w_i$ of duration $\Delta t=30\,\text{s}$
defined in Section~\ref{sec:seem-stage-1}.
Using the main-text notation for eligible records in this retrieval stage,
\begin{equation}
    \artifacts_q =
    \{r\in\mathcal{M}_{\text{text}}:
      r.t_{\text{end}}\leq t_q\}.
    \label{eq:eligible-event-bank}
\end{equation}
The comparison uses the window's end time: a window that starts before
$t_q$ but ends after it is excluded.
Both ranking and support counting operate on $\artifacts_q$, so future
records cannot influence either the displayed evidence or its support
statistics.

\noindent\textbf{Field-aware question matching.}
Let $T(x)$ denote the lexical token set of text $x$.
The field set $\mathcal{K}$ consists of \texttt{EVENT}, \texttt{PEOPLE},
\texttt{OBJECTS}, \texttt{PLACE}, and \texttt{DIALOGUE};
it is distinct from the policy frontier $\frontier_k$.
For $\phi\in\mathcal{K}$, let $r_\phi$ denote that field's text and define
$s_\phi(r)=|T(q)\cap T(r_\phi)|$, as in the main text.
Matching is performed separately within each field.
The base question score is
\begin{equation}
\begin{aligned}
    S_q^{\mathrm{base}}(r) ={}&
    3.0\,s_{\texttt{EVENT}}(r)
    +1.5\,s_{\texttt{PEOPLE}}(r)
    +1.5\,s_{\texttt{OBJECTS}}(r)\\
    &+1.25\,s_{\texttt{PLACE}}(r)
    +2.0\,s_{\texttt{DIALOGUE}}(r).
\end{aligned}
    \label{eq:fielded-question-base}
\end{equation}
Table~\ref{tab:fielded-question-weights} lists these coefficients.
The score favors agreement with the event action while retaining
entity, location, and speech evidence.
Question-form adjustments further emphasize \texttt{PLACE} for
``where'' questions, \texttt{PEOPLE} for ``who'' or ``with whom''
questions, and \texttt{DIALOGUE} for speech or conversation questions.
We denote the resulting question score by
$S_q(r)=\sum_{\phi\in\mathcal{K}}w_\phi(q)s_\phi(r)$.
The table gives the base weights $w_\phi^{\mathrm{base}}$;
$w_\phi(q)$ includes the question-form adjustments.
These coefficients are fixed retrieval rules in the final policy,
selected during \method{}'s policy search.

\begin{table}[htbp]
\centering
\caption{Base question-matching weights in fielded event retrieval.
Each coefficient multiplies the lexical overlap with its field.}
\label{tab:fielded-question-weights}
\begin{tabular}{lccccc}
\toprule
Field & \texttt{EVENT} & \texttt{PEOPLE} & \texttt{OBJECTS}
      & \texttt{PLACE} & \texttt{DIALOGUE} \\
\midrule
$w_\phi^{\mathrm{base}}$ & $3.0$ & $1.5$ & $1.5$ & $1.25$ & $2.0$ \\
\bottomrule
\end{tabular}
\end{table}

\noindent\textbf{Choice matching and same-event support.}
For every answer choice $c\in A_q$, the policy separately computes
$S_c(r,c)$ from the choice's overlap with the record's fields.
People matches receive weight $3.5$, and an exact choice-phrase match
adds $5.0$ to the choice score.
With the same $S_q(r)$, $S_c(r,c)$, and coefficient meanings as
Eq.~\ref{eq:event-retrieval-score}, question and choice relevance combine as
\begin{equation}
    S(r,c)=
    S_q(r)+2.0\,S_c(r,c)
    +4.0\,\mathbb{I}
    [S_q(r)>0\ \wedge\ S_c(r,c)>0].
    \label{eq:fielded-retrieval-score}
\end{equation}
Thus $\alpha=1$ multiplies $S_q$, $\beta=2$ multiplies $S_c$, and
$\gamma=4$ is the same-record co-occurrence bonus, exactly as in the main text.
The $4.0$ bonus is awarded once for each record and choice pair when both
scores are positive.
It promotes evidence where the question and choice are mentioned in
the same event window; lexical co-occurrence alone does not establish
that the record entails the answer.
For example, if $S_q(r)=3.0$ and $S_c(r,c)=3.5$, the combined score
is $3.0+2.0\times3.5+4.0=14.0$ before temporal reranking.
The $5.0$ exact-phrase bonus in $S_c$ adds $10.0$ to the total.

\noindent\textbf{Temporal ordering and support counts.}
After lexical scoring, a positional bias favors later records for
``last,'' ``latest,'' or ``previous'' queries and earlier records for
``first'' or ``earliest'' queries.
This reranking operates within the eligible bank and does not change
the cutoff in Eq.~\ref{eq:eligible-event-bank}.
The policy also scans the eligible bank to compute
\begin{equation}
    n_c=\sum_{r\in\artifacts_q}
    \mathbb{I}[S_q(r)>0\ \wedge\ S_c(r,c)>0].
    \label{eq:fielded-support-count}
\end{equation}
Each qualifying record contributes exactly $1$ to $n_c$, irrespective
of its ranking score; one record may contribute to multiple choices.
The count covers all eligible records, including records omitted from
the packed context.
For habit questions, repeated support across timestamps supplies
evidence beyond a single salient event.
The structured fallback stores these statistics as
\texttt{choice\_support\_counts} and renders them as
\texttt{field\_cooccurrence\_count}.
They are lexical support counts, not calibrated answer probabilities,
and differ from the caption-level \texttt{choice\_hit\_counts} used by
the visual gate.

\noindent\textbf{Evidence packing and answer generation.}
The structured context $c_q$ contains up to $6$ globally relevant event
anchors and up to $3$ evidence records for each answer choice.
With $4$ choices, this allocates at most $6+4\times3=18$ record
slots before the character cap; the slots need not represent $18$
distinct events.
The context is capped at $12{,}500$ characters.
It includes the question type, memory mode, temporal hint, number
of eligible records, per-choice support counts, and each selected
record's score, timestamp, and available semantic fields.
The responder is instructed to check action agreement in
\texttt{EVENT}, require same-record evidence for people and places,
prioritize \texttt{DIALOGUE} for speech questions, and aggregate
repeated evidence for habits.
Global mention frequency alone is insufficient to select an answer.
When the structured fallback is used, one text-model call resolves
the packed evidence into a single answer letter from A to D.

\subsection{Evidence Specialist Rules}
\label{app:specialists}

The specialists run in the following order. Each requires evidence suited
to its operation and abstains when that requirement is unmet.

\noindent\textbf{Temporal Delta Specialists.}
For relative-time queries such as ``When was the last time\ldots?'',
the specialist localizes the most recent matching event anchor, computes
the time delta relative to $t_q$, and maps the duration to the answer
choices, answering without an LLM call.

\noindent\textbf{Frequency Aggregators.}
For recurring behavior queries, the specialist aggregates support counts
across distinct timestamps. It answers directly only when the leading
choice has a clear support margin over the alternatives.

\noindent\textbf{Relational and Provenance Specialists.}
For queries about interactions, such as meal companions or joint tasks,
the specialist binds entities within the same event: all queried entities
must co-occur in the same action record before it can answer.
Unrelated records cannot establish this association.

\noindent\textbf{Structured Event Fallback.}
If all specialists abstain, the packed event context is passed to the
fixed model $f$ for reasoning across the structured fields:
$\hat{a}_t=f(q,A_q,c_q)$.
Here the subscript $t$ in $\hat{a}_t$ and $z_t$ denotes the text stage;
the query cutoff remains $t_q$.
The resulting text answer $\hat{a}_t$ and route metadata $z_t$ are passed
to the visual gate, as are answers produced directly by specialists.

\subsection{Visual Gate and Verification}
\label{app:visual-verification}

The gate in Eq.~\ref{eq:visual-gate} requires both a visual need and weak
text evidence.

\noindent\textbf{Visual-Need Predicate.}
$\mathrm{StrongVisualNeed}(q)$ is a deterministic lexical rule applied to
the concatenation of the question-type label and question text.
It does not inspect the answer choices, text-stage answer, retrieved
evidence, or frame availability, and requires no model call.
Matching is case-insensitive.
At least one term in Table~\ref{tab:visual-need-triggers} must match.

\begin{table}[htbp]
    \centering
    \caption{Examples of trigger vocabulary for $\mathrm{StrongVisualNeed}(q)$.
    A trigger match is necessary but can be overridden by an exclusion rule.}
    \label{tab:visual-need-triggers}
    \small
    \begin{tabularx}{\linewidth}{lX}
        \toprule
        Category & Trigger terms \\
        \midrule
        Color & \texttt{color}, \texttt{colour} \\
        Counting & \texttt{how many}, \texttt{number of}, \texttt{count} \\
        Position & \texttt{left}, \texttt{right}, \texttt{above}, \texttt{below} \\
        Text or screen & \texttt{sign}, \texttt{text}, \texttt{word}, \texttt{screen} \\
        Visible state & \texttt{visible}, \texttt{shown}, \texttt{look}, \texttt{looks} \\
        Worn or held objects & \texttt{wearing}, \texttt{wear}, \texttt{held}, \texttt{holding} \\
        Shape or markings & \texttt{shape}, \texttt{brand}, \texttt{logo} \\
        \bottomrule
    \end{tabularx}
\end{table}

Five exclusion rules then suppress matches associated with temporal
aggregation, identity, preference, or relational wording.
Let $\mathrm{Trigger}(q)$ denote a trigger match and $E_j(q)$ an exclusion match.
The predicate is
\begin{equation}
    \mathrm{StrongVisualNeed}(q)
    = \mathrm{Trigger}(q) \wedge \neg\left(\bigvee_{j=1}^{5} E_j(q)\right).
    \label{eq:strong-visual-need}
\end{equation}
The exclusions are:
\begin{enumerate}
    \item \textbf{Temporal frequency:}
    \texttt{how many times}, \texttt{number of times}, \texttt{most often},
    \texttt{past N days} (with numeric $N$), \texttt{usually},
    \texttt{often}, \texttt{frequency}, or \texttt{frequently}.
    \item \textbf{Unconstrained identity:}
    questions containing \texttt{who}, unless it is immediately followed
    by \texttt{is}, \texttt{was}, \texttt{were}, \texttt{appear(s)},
    \texttt{look(s)}, \texttt{wearing}, \texttt{wears}, \texttt{holding},
    \texttt{holds}, \texttt{visible}, or \texttt{shown}, with a subsequent
    cue from \texttt{wearing}, \texttt{wears}, \texttt{holding},
    \texttt{holds}, \texttt{visible}, \texttt{shown}, \texttt{color},
    or \texttt{colour}.
    Here \texttt{appear(s)} and \texttt{look(s)} denote the base
    and third-person singular forms.
    \item \textbf{Preference-related wording:}
    \texttt{like}, \texttt{likes}, \texttt{prefer}, \texttt{prefers},
    \texttt{drink}, or \texttt{drinks}.
    \item \textbf{Nonvisual idiom:} the phrase \texttt{look after}.
    \item \textbf{Position with relation or negation:}
    a positional term (\texttt{left}, \texttt{right}, \texttt{above},
    or \texttt{below}) together with \texttt{who}, \texttt{whom},
    \texttt{whose}, \texttt{with}, \texttt{together}, \texttt{not},
    \texttt{n't}, \texttt{did not}, \texttt{without}, or \texttt{participate}.
\end{enumerate}
For example, ``Who is wearing the red shirt?'' satisfies the visual-need
predicate, whereas ``How many times did I wear the red shirt?'' is
excluded as a temporal-frequency query.
These rules approximate visual need from wording; they do not establish
that the retrieved frames can answer the question.
They can also miss visual queries: the preference exclusion suppresses
\texttt{look like}, and unlisted inflections do not trigger a match.

\noindent\textbf{Text-Evidence Check and Verification.}
For $\mathrm{WeakText}(\hat{a}_t,z_t)$, let $p$ indicate whether the text
answer is parseable, $d$ whether the route was direct, $\rho_t$ the route name,
and $h_1,h_2$ the top two lexical choice-hit counts.
The route name $\rho_t$, direct-route flag $d$, and counts are read from
$z_t$; $r$ remains reserved for an event record.
The text evidence is considered weak unless all reliability checks hold:
\begin{equation}
    \mathrm{WeakText}(\hat{a}_t, z_t) = \neg \Big( p \;\wedge\; d \;\wedge\; (\texttt{"direct"} \in \rho_t) \;\wedge\; (h_1 > 2) \;\wedge\; (h_1 - h_2 > 1) \Big).
    \label{eq:weak-text}
\end{equation}
When the gate remains closed, the policy returns $\hat{a}_t$ without
visual verification, so the final prediction in
Section~\ref{sec:problem-formulation} is $\hat y_q=\hat{a}_t$.
Otherwise, it ranks caption windows and retrieves at most four relevant
frames ($N_{\text{frames}} \le 4$) from $\mathcal{M}_{\text{vis}}$.
The fixed model $f$ inspects these localized frames and is instructed to
retain the text answer unless visual evidence clearly supports an alternative.
Missing frames or verification failures also leave $\hat{a}_t$ unchanged.

\noindent\textbf{Conservative Visual Verification Prompt.}
\label{app:prompts}
When the visual gate opens, the multimodal verifier is guided by the following prompt:
\begin{quote}
\emph{``Current text answer is $\hat{a}_t$. Inspect frames. Maintain $\hat{a}_t$ unless visual evidence unambiguously proves an alternative.''}
\end{quote}

\section{Additional Results}
\label{app:additional-results}


\subsection{Visual Gate Design}
\label{app:visual-gate-ablation}

Table~\ref{tab:ablation_visual_gate} compares the evolved rule-based gate
with an LLM judge on EgoLifeQA, Ego-R1 Bench, and Video-MME (L).
The rule-based gate invokes visual verification $112$ times, compared
with $450$ for the LLM judge, or approximately $4.0\times$ fewer calls.

\begin{table}[htbp]
\caption{Visual-gate ablation with GPT-5 mini across EgoLifeQA,
Ego-R1 Bench, and Video-MME (L). Accuracy is reported in percent.
Calls reports visual-verification calls
across the three benchmarks, excluding LLM gate-decision calls.}
\label{tab:ablation_visual_gate}
\centering
\begin{tabular}{lcccc}
\toprule
Gate & EgoLife & Ego-R1 & VMME-L & Calls \\
\midrule
LLM judge      & 53.0 & 52.7 & 55.7 & 450 \\
Evolved rules (\memoryshort{}) & 53.6 & 53.0 & 55.9 & \textbf{112} \\
\bottomrule
\end{tabular}
\end{table}

\subsection{Answer-Time Cost with Open Backbones}
\label{app:efficiency-open}

\noindent \textbf{Deployment cost with an open backbone.}
Table~\ref{tab:efficiency-open}
reports per-question cost when the answering model is the open Qwen3-VL-8B
checkpoint served locally, with memory built by GPT-5 mini as everywhere else in
this paper.  Across the three benchmarks, mean per-question cost remains below
$5.7$k tokens and $5$\,s, with at most $1.25$ model requests per question on
average.  On the two
personal-memory benchmarks, the request count is below one ($0.94$ on
EgoLifeQA and $0.96$ on Ego-R1 Bench) because a fraction of questions is
resolved from memory without a model call at all, a floor that a per-question
frame-prompting interface cannot reach.

\begin{table}[htbp]
\caption{Answer-time cost of \memoryshort{} with Qwen3-VL-8B as the answering
model.  Tok./q and Req./q are tokens and model requests per question, Wall/q is
end-to-end per-question latency, and LLM is the part of that latency spent
inside the model call. $N_{\mathrm{QA}}$ is the number of evaluated questions.
Questions are answered serially within a run, so Wall/q
is measured per-question latency rather than throughput divided by question
count.}
\label{tab:efficiency-open}
\centering
\begin{tabular}{lrrrrr}
\toprule
Benchmark & $N_{\mathrm{QA}}$ & Tok./q & Req./q & Wall/q (s) & LLM (s, \%) \\
\midrule
EgoLifeQA & 500 & 3459 & 0.94 & 2.24 & 1.10 (49\%) \\
Ego-R1 Bench & 300 & 3473 & 0.96 & 1.85 & 0.66 (36\%) \\
Video-MME (L) & 900 & 5630 & 1.25 & 4.99 & 1.54 (31\%) \\
\bottomrule
\end{tabular}

\end{table}

\noindent \textbf{Where the time goes.}
The last column of Table~\ref{tab:efficiency-open} shows that with a locally
served 8B backbone, the model call accounts for $31\%$ to $49\%$ of end-to-end
latency; the rest is memory retrieval, frame extraction, and encoding.
This breakdown identifies substantial local processing overhead alongside
model inference. Table~\ref{tab:ablation-latency} reports total per-question
evaluation costs for a different answering configuration without decomposing
API time, so its wall times cannot establish the same breakdown.

\noindent \textbf{The modality gate makes latency bimodal.}
Video-MME (L) latency is bimodal: the median is $0.71$\,s, the mean $4.99$\,s,
and the p95 $20.4$\,s.
The $675$ text-only questions average $0.98$\,s and $3362$ tokens, with a
$0.70$\,s median; the $225$ gated questions average $17.00$\,s and $12433$
tokens, with a $14.66$\,s median and $28.4$\,s p95.
Visual escalation thus costs roughly $17\times$ the time and $3.7\times$ the
tokens on one quarter of the questions, making the overall mean unrepresentative
of either route.
EgoLifeQA and Ego-R1 Bench have narrow latency distributions and p95 near $3.5$\,s.

\noindent \textbf{Comparison with direct frame prompting.}
Direct prompting of Qwen3-VL-8B with uniformly sampled frames costs $7153$
tokens and $6.65$\,s per question on EgoLifeQA, $1317$ tokens and $1.36$\,s on
Ego-R1 Bench, and $4684$ tokens and $3.63$\,s on Video-MME (L).
On EgoLifeQA, \memoryshort{} uses $2.1\times$ fewer tokens and $3.0\times$
less wall time by retrieving event-anchor text instead of processing frames
from a full day.
Memory costs more on Ego-R1 Bench ($2.6\times$ the tokens and $1.4\times$
the time) and Video-MME (L) ($1.2\times$ and $1.4\times$, respectively).
The cost advantage is clearest for repeated queries over long egocentric
recordings, where direct prompting re-encodes frames for each question.

\subsection{Cost Accounting: Memory Construction and Evaluation}
\label{app:memory-cost}

Table~\ref{tab:memory_construction_cost} separates offline construction from
$500$-question EgoLifeQA evaluation with GPT-5 mini.
\memoryshort{} constructs memory from $30$-second captions and event-anchor
rewrites.
WorldMM's evaluation includes index construction and question answering;
indexing is distinct from offline construction and counted once.
Combined construction and evaluation use $18.72$M tokens for \memoryshort{}
and $36.77$M for WorldMM, excluding policy-search costs.

\begin{table}[htbp]
\caption{Offline construction, evaluation, and combined cost comparison between WorldMM and \memoryshort{} on EgoLifeQA
($500$ questions) with GPT-5 mini.}
\label{tab:memory_construction_cost}

\centering
\begin{tabular}{llrrr}
\toprule
\textbf{Phase} & \textbf{Method} & \textbf{Requests} & \textbf{Total Tokens} & \textbf{Time (s)} \\
\midrule
Offline construction & WorldMM & 28,900 & 25.15M & 13,320 \\
 & \memoryshort{} & 12,446 & 17.08M & 2,541 \\
\midrule
Evaluation ($500$ questions) & WorldMM & 15,233 & 11.62M & 61,594 \\
 & \memoryshort{} & 485 & 1.647M & 1,375 \\
\midrule
Total & WorldMM & 44,133 & 36.77M & 74,914 \\
 & \memoryshort{} & 12,931 & 18.72M & 3,916 \\
\bottomrule
\end{tabular}

\end{table}

By deferring relational reasoning to retrieval over structured text,
\memoryshort{} removes offline information extraction and semantic graph
consolidation, more than halving per-window model calls
(Table~\ref{tab:pipeline_comparison}).

\begin{table}[htbp]
\caption{Comparison of memory construction complexity per 30-second video window.}
\label{tab:pipeline_comparison}
\centering
\begin{tabularx}{\linewidth}{lXr}
\toprule
\textbf{Method} & \textbf{Operations per 30s window} & \textbf{Calls} \\
\midrule
WorldMM & Caption (1) + OpenIE (2) + Semantic Ext. (0.1) + Consolidation (1.45) & $\sim$4.6 \\
\textbf{\memoryshort{} (Ours)} & Caption (1) + Event-Anchor Rewriting (1) & \textbf{2.0} \\
\bottomrule
\end{tabularx}

\end{table}

%
%

\section{Limitations and Discussion}
\label{app:limitations}

Despite its efficiency at test time, SEEM currently relies on offline caption generation for initial memory ingestion, and its search was bounded by modular policy templates. In future work, we plan to extend LVMH towards online streaming memory construction for continuous real-time video streams, expand the harness search space to discover dynamic multi-scale hierarchical graphs, and deploy the evolved policies to physical embodied robotic platforms.

Although \method{} does not fully resolve development-set overfitting,
our experiments and ablation study show better generalization than
Meta-Harness and the single-agent baseline in the reported runs.
An important contribution of this work is to highlight development-set
overfitting in automated harness evolution, emphasize its impact on
policy search, and introduce a framework that improves generalization
in our evaluations.
The information available to a coding agent, including task
examples, execution traces, and evaluation feedback, can strongly shape
its hypotheses, evolution trajectory, and resulting policies.
Restricting direct access to evaluation examples therefore does not by
itself ensure generalization.
At a conceptual level, harness self-evolution by coding agents resembles
neural-network optimization: evaluation feedback serves as an objective
signal analogous to a loss, while the agent interprets its observations
to propose update directions, rather than computing gradients.
This perspective makes the design of agent observations an important
part of the optimization process.
We view understanding this process as an important research direction
and plan to systematically study how the content and granularity of
observations affect policy evolution, overfitting, and generalization,
with the goal of developing harnesses that generalize more reliably.

\end{document}